\documentclass{article}
\usepackage{ijcai26}

\usepackage{times}
\usepackage{soul}
\usepackage{url}
\usepackage[hidelinks]{hyperref}
\usepackage[utf8]{inputenc}
\usepackage[small]{caption}
\usepackage{graphicx}
\usepackage{amsmath}
\usepackage{amsthm}
\usepackage{booktabs}
\usepackage{algorithm}
\usepackage{algorithmic}
\usepackage[switch]{lineno}

\usepackage{amsfonts}
\usepackage{multirow}
\usepackage{array}
\usepackage{subcaption}
\usepackage{float}
\usepackage{xcolor} % 支持颜色命令
\usepackage{colortbl} % 支持表格列底色

\title{Explicit Uncertainty Modeling for Active CLIP Adaptation with Dual Prompt Tuning}

\author{
Qian-Wei Wang$^{1,2}$\and
Yaguang Song$^2$\And
Shu-Tao Xia$^{1,2}$ \\
\affiliations
$^1$Tsinghua Shenzhen International Graduate School, Tsinghua University\\
$^2$Institute of Perceptual Intelligence, Peng Cheng Laboratory\\
\emails
wanggw21@mails.tsinghua.edu.cn,
songyg01@pcl.ac.cn,
xiast@sz.tsinghua.edu.cn
}

\begin{document}

\maketitle

\begin{abstract}
Pre-trained vision–language models such as CLIP exhibit strong transferability, yet adapting them to downstream image classification tasks under limited annotation budgets remains challenging. In active learning settings, the model must select the most informative samples for annotation from a large pool of unlabeled data. Existing approaches typically estimate uncertainty via entropy-based criteria or representation clustering, without explicitly modeling uncertainty from the model perspective. In this work, we propose a robust uncertainty modeling framework for active CLIP adaptation based on dual-prompt tuning. We introduce two learnable prompts in the textual branch of CLIP. The positive prompt enhances the discriminability of task-specific textual embeddings corresponding to light-weight tuned visual embeddings, improving classification reliability. Meanwhile, the negative prompt is trained in an reversed manner to explicitly model the probability that the predicted label is correct, providing a principled uncertainty signal for guiding active sample selection. Extensive experiments across different fine-tuning paradigms demonstrate that our method consistently outperforms existing active learning methods under the same annotation budget.
\end{abstract}

\section{Introduction}

Large-scale pre-trained vision–language models (VLMs) \cite{radford2021learning,bai2025qwen2}, such as CLIP, have emerged as highly effective foundation models for visual understanding. Trained on massive collections of image–text pairs with contrastive objectives, these models learn rich joint representations that can generalize across diverse visual tasks without requiring extensive task-specific supervision. However, despite these impressive zero-shot capabilities, VLMs are not immune to domain-specific challenges. When transferred to specialized downstream tasks, such as satellite remote sensing image classification, that exhibit significant domain divergence from the pre-training distribution, the performance of zero-shot inference can be substantially lower than desired. This observation highlights that, although VLMs encode strong generic visual–semantic priors, they still require efficient adaptation to realize their full potential on domain-specific tasks. 

Collecting labeled data for downstream tasks, however, is costly and labor-intensive. Active learning (AL) addresses this challenge by iteratively selecting the most informative samples from a large pool of unlabeled data for annotation, thereby reducing annotation costs while improving model performance. For example, the classification accuracy on EuroSAT dataset can increase from 42.0\% to 91.2\% with only 1\% of selected samples with human annotated labels. Despite the recent progress in combining active learning with VLMs, current approaches exhibit several limitations in how they estimate sample uncertainty and guide annotation. Most works rely on heuristic uncertainty measures such as predictive entropy, margin scores, or representation clustering, or design complex calibration schemes that fuse multiple proxy signals to approximate informativeness. Other approaches focus on addressing dataset-level issues such as class imbalance through re-weighting or sampling adjustments. These strategies treat the model as a black box and build uncertainty estimators on top of its outputs or intermediate embeddings. 

These limitations motivate the need for an AL framework that can explicitly and robustly estimate per-sample uncertainty by leveraging the structural properties of the vision–language model itself, rather than relying on post-hoc or heuristic criteria. In this paper, we propose a simple yet effective uncertainty modeling framework for active CLIP adaptation based on dual prompt tuning. Our method introduces two classes of learnable prompts in the textual branch of CLIP: a positive prompt that enhances class discriminability by aligning lightweight visual embeddings with task-specific textual representations, and a negative prompt trained in a reversed manner to explicitly capture the likelihood that a pseudo-label is correct. This dual mechanism enables robust estimation of clean probabilities for pseudo-labels, which we then use to rank samples within each predicted class to guide both uncertainty-aware query selection and confident pseudo-label mining during AL iterations.

In experiments, our method consistently achieves substantial performance gains across different datasets, annotation budgets, and fine-tuning paradigms, demonstrating that our model-integrated design, built from the perspective of architectural innovation rather than relying on heuristic or post-hoc strategies based solely on model outputs, effectively enhances sample selection and uncertainty estimation.

\section{Related Work}

\subsection{Pre-trained VLMs and Parameter-Efficient Fine-Tuning}
The emergence of large-scale pre-trained vision-language models (VLMs) has revolutionized the field of visual understanding by bridging the gap between visual and textual modalities. Unlike single-modality pre-trained models that focus solely on visual features or linguistic representations, VLMs learn joint vision-language embeddings through contrastive learning on massive image-text pairs, enabling strong zero-shot and few-shot transfer capabilities. Radford et al. \cite{radford2021learning} pioneered this direction with CLIP, which aligns visual encoders (ViT or ResNet) and textual encoders (Transformer) via a contrastive loss over millions of image-text pairs. Subsequent VLMs, such as ALIGN \cite{jia2021scaling}, FLAVA \cite{singh2022flava}, and Qwen-VL \cite{bai2025qwen2}, further improve performance by expanding pre-training data scales, optimizing architecture designs, or enhancing cross-modal alignment, solidifying VLMs as foundational models for diverse downstream tasks.

Despite their impressive generalization, VLMs often require task-specific adaptation to achieve optimal performance on domain-specific downstream tasks, especially when facing distribution shifts. Full fine-tuning, which updates all model parameters, suffers from high computational costs and risks overfitting under limited annotation budgets. This has spurred the development of parameter-efficient fine-tuning (PEFT) methods, which only optimize a small subset of parameters while freezing the pre-trained backbone, striking a balance between adaptation effectiveness and computational efficiency.

Prompt learning, a dominant PEFT paradigm for VLMs, modifies the input space of encoders to guide the model toward task-specific objectives without altering the pre-trained weights. For textual prompts, early works such as CoOp \cite{zhou2022learning} and CoCoOp \cite{zhou2022conditional} introduce learnable continuous prompt tokens for the textual branch of CLIP, adapting the model to image classification tasks by optimizing task-aware textual representations. These methods demonstrate that well-learned textual prompts can significantly enhance the discriminability of class embeddings, outperforming zero-shot inference and even full fine-tuning in low-data regimes. Extensions of textual prompt learning include hybrid prompt strategies that combine handcrafted and learnable tokens \cite{cao2024adaclip} and dynamic prompt generation that adjusts prompts based on input samples \cite{he2025dss}.

Complementary to textual prompt learning, visual prompt tuning (VPT) extends PEFT to the visual branch of VLMs, addressing scenarios where domain-specific visual features deviate significantly from pre-training distributions. Jia et al. \cite{jia2022visual} first proposed VPT, which prepends a small set of learnable visual prompt tokens to the input sequence of the visual transformer, enabling fine-grained adaptation of visual embeddings without modifying the pre-trained visual backbone. Subsequent works have enhanced VPT by optimizing prompt positions \cite{tang2025visual}, incorporating hierarchical prompts \cite{zheng2025hierarchical}, or combining visual prompts with textual prompts for cross-modal alignment. Recent advances in visual prompt learning, such as progressive prompt tuning \cite{xu2025progressive}, further improve robustness by incrementally adjusting prompt complexity during training, which is particularly valuable for limited-data settings.

While existing prompt learning methods primarily focus on enhancing classification accuracy rather than modeling uncertainty. Our work differs by introducing dual-prompt that not only align visual and textual embeddings but also explicitly capture pseudo-label reliability, enabling robust uncertainty estimation for active learning.

\begin{figure*}
    \centering
    \includegraphics[width=0.85\linewidth]{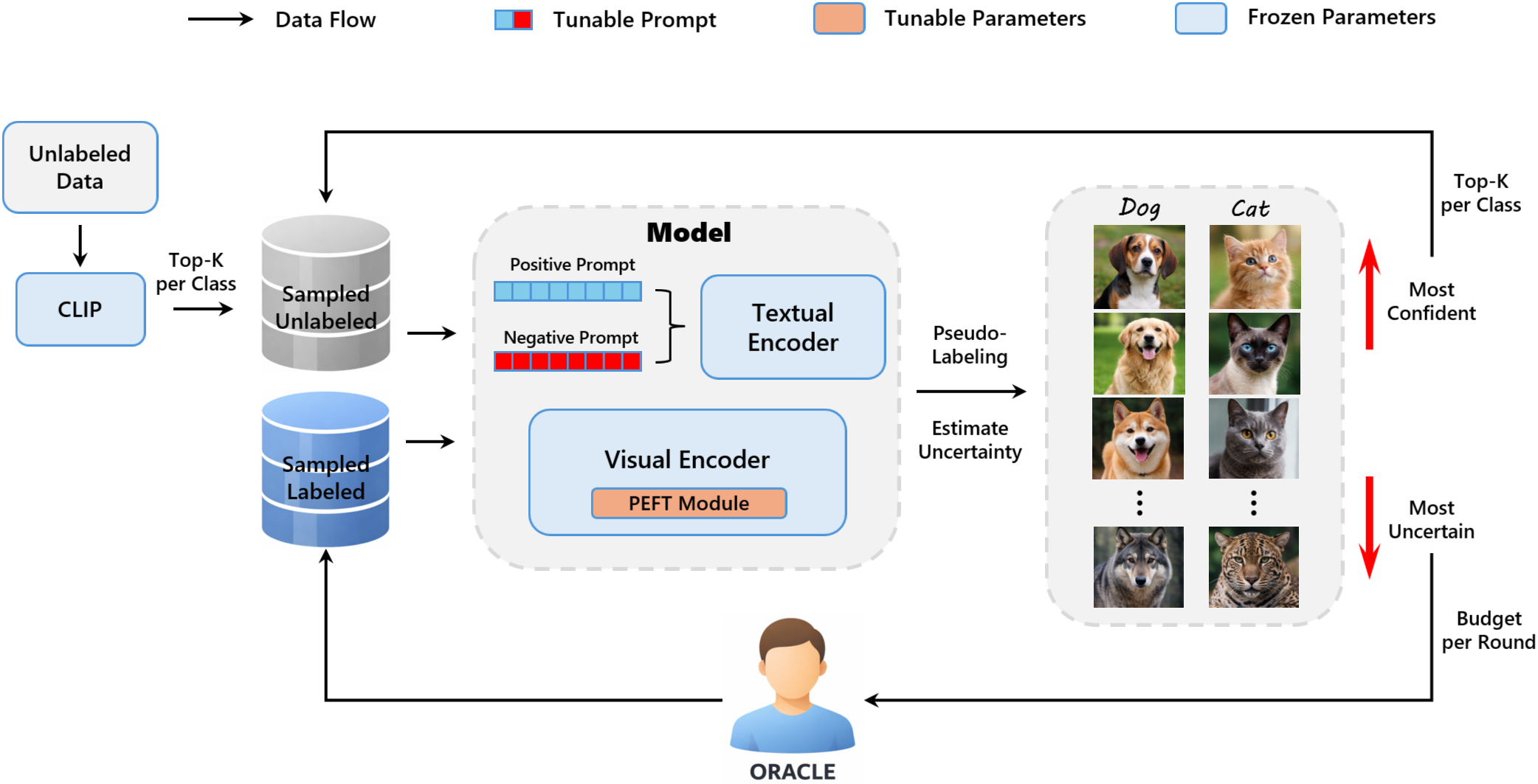}
    \caption{The overall illustration of our method.}
    \label{fig:overall}
\end{figure*}

\subsection{Active Learning for VLMs}

Active learning (AL) addresses the challenge of costly label acquisition by iteratively selecting the most informative unlabeled samples for annotation, guiding model training with minimal human supervision \cite{settles2009active}. The core of AL lies in designing effective query strategies that identify samples which maximize model performance gains, with three dominant paradigms: uncertainty sampling, diversity sampling, and hybrid strategies that combine both \cite{ren2021survey}.

Early AL works for computer vision focus on traditional CNN models and rely on heuristic uncertainty measures. Uncertainty sampling methods select samples where the model exhibits the highest ambiguity, such as those with maximum entropy \cite{qiu2016maximum}, minimum margin between top predictions \cite{balcan2007margin}, or maximum disagreement among an ensemble of models \cite{di2011view}. Diversity sampling, on the other hand, ensures selected samples cover the data distribution to avoid redundancy, often using clustering \cite{tuia2010cluster} or distance metrics \cite{wu2021redal} to select representative samples. Hybrid strategies, such as CoreSet \cite{sener2017active}, integrate uncertainty and diversity by selecting samples that are both uncertain and representative of the unlabeled pool, becoming a de facto baseline for AL in computer vision.

With the rise of VLMs, recent works have explored AL for adapting pre-trained VLMs to downstream tasks, leveraging the strong zero-shot capabilities of VLMs to initialize the AL pipeline and design AL strategies to guide sample selection for adaptation. 

Some researchers \cite{safaei2025active} observed that directly applying traditional uncertainty measures, such as predictive entropy or margin scores, to VLM outputs can lead to unreliable sampling decisions due to mis-calibrated predictions. To address this, they extended uncertainty estimation to cross-modal embeddings and combined self-uncertainty with neighborhood-aware measures, providing more stable sample selection and improving adaptation performance across multiple datasets. In another study \cite{bang2024active}, the authors highlighted that naively integrating AL with VLMs often underperforms random sampling at early stages, largely because of imbalanced distributions among unlabeled candidates, which may lead to biased selection of labeled samples. To mitigate this issue, they proposed pre-balancing the candidate set based on internal VLM representations, thereby improving the class coverage of selected samples and enhancing the effectiveness of AL-driven adaptation. Furthermore, some researchers \cite{wang2025active} explored leveraging open pre-training data to augment downstream tasks. By retrieving samples from pre-training corpus that are relevant to target classes, and prioritizing high-uncertainty samples from underrepresented categories through a tail first sampling strategy, ALOR demonstrated that combining retrieved data with uncertainty-aware sampling can substantially improve AL performance, especially under limited annotation budgets.

Despite these advances, these methods generally rely on post-hoc processing of model outputs, uncertainty calibration, or external data augmentation to estimate sample informativeness, rather than designing model-integrated structure of the VLM to uncertainty uncertainty.

\section{Methodology}

\subsection{Overview}

We study an \textbf{active learning (AL)} setting for adapting large pre-trained vision–language models (VLMs), specifically CLIP, to downstream image classification tasks under limited annotation budgets. At each AL round, the model selectively queries the most informative samples for human annotation while leveraging confident pseudo-labeled samples to improve data efficiency.

To this end, we propose an uncertainty modeling framework for active CLIP adaptation with \emph{Dual Prompt Tuning}. Our approach (See Fig.\ref{fig:overall}) consists of two tightly coupled components:

\begin{enumerate}
    \item A \textbf{dual-prompt-based CLIP adaptation model}, which robustly estimates the clean probability of pseudo-labels via positive and negative textual learnable prompts, together with parameter-efficient adaptation of the visual encoder.
    \item An \textbf{uncertainty-driven active learning strategy}, which embeds the proposed model into a round-based AL loop for uncertainty-aware query selection and confident pseudo-label mining.
\end{enumerate}

We next describe the two components in detail.

\subsection{Dual-Prompt-Based CLIP Adaptation Model}

\subsubsection{Textual Dual Prompt Learning}

Given a downstream classification task with \(C\) categories, we adapt a pre-trained CLIP model by introducing \textbf{dual prompt learning} in the textual encoder to robustly estimate the clean probability of pseudo-labels.

For each class \(k \in \{1, \dots, C\}\), we maintain two learnable prompts: a \textbf{positive prompt} and a \textbf{negative prompt}, defined as
\begin{align}
    \text{prompt}^+_k &= [V^+_1], [V^+_2], \dots, [V^+_M], \text{[CLS]}_k, \\
    \text{prompt}^-_k &= [V^-_1], [V^-_2], \dots, [V^-_M], \text{[CLS]}_k,
\end{align}
where \(V^+_i, V^-_i \in \mathbb{R}^d\) are learnable context tokens, \(M\) is the number of tunable textual tokens, and \(\text{[CLS]}_k\) denotes the class-specific token corresponding to category \(k\).

The positive and negative prompts are jointly optimized to provide a robust estimate of pseudo-label reliability. Specifically, for a sample \(\boldsymbol{x}\) with pseudo-label \(\hat{y}\), the clean probability is defined as:
\begin{equation}
    \tiny
    \label{pclean}
    p^{\text{clean}}_{\hat{y}} =
    \frac{
        \exp\left(\text{sim}\left(v_{\theta_v}(\boldsymbol{x}), t_{\theta_t}^+(\boldsymbol{c}_{\hat{y}})\right)/\tau\right)
    }{
        \exp\left(\text{sim}\left(v_{\theta_v}(\boldsymbol{x}), t_{\theta_t}^+(\boldsymbol{c}_{\hat{y}})\right)/\tau\right)
        +
        \exp\left(\text{sim}\left(v_{\theta_v}(\boldsymbol{x}), t_{\theta_t}^-(\boldsymbol{c}_{\hat{y}})\right)/\tau\right)
    },
\end{equation}
where \(\text{sim}(\cdot, \cdot)\) denotes cosine similarity, \(\tau\) is a temperature hyperparameter, \(v_{\theta_v}(\cdot)\) is the visual embedding modulated by visual prompts, and \(t_{\theta_t}^+(\cdot), t_{\theta_t}^-(\cdot)\) are the positive and negative textual embeddings, respectively.

This design provides a per-sample estimation of pseudo-label uncertainty, which is subsequently used to rank samples within each class to select the most uncertain and  most confident instances for annotation and training.

To train the dual prompts, we define the following two losses:
\begin{itemize}
    \item Loss for positive prompts and visual prompts (\(\mathcal{L}_1\)): This aligns visual embeddings with their corresponding positive textual embeddings. For a sample \(\boldsymbol{x}\) with human-annotated or pseudo-label \(\hat{y}\), we minimize the cross-entropy between the predicted distribution and the pseudo-label:
    \begin{equation}
    \small
        \mathcal{L}_1 = - \log
        \frac{\exp(\text{sim}(v_{\theta_v}(\boldsymbol{x}), t_{\theta_t}^+(\boldsymbol{c}_{\hat{y}})) / \tau)}{\sum_{k=1}^C \exp(\text{sim}(v_{\theta_v}(\boldsymbol{x}), t_{\theta_t}^+(\boldsymbol{c}_k)) / \tau)}.
    \end{equation}
    
    \item Loss for negative prompts (\(\mathcal{L}_2\)): This encourages positive prompts to be more discriminative than negative prompts for true labels, while reversing the relation for noisy labels. For each sample \(\boldsymbol{x}\) with human-annotated or pseudo-label \(\hat{y}\), we randomly sample a complement label \(\tilde{y}\) from other classes to simulate noise. With \(p^{\text{clean}}_{\hat{y}}\) from Eq.\ref{pclean}, we define:
    \begin{equation}
        \mathcal{L}_2 = - \log p^{\text{clean}}_{\hat{y}} + \log (1 - p^{\text{clean}}_{\tilde{y}}).
    \end{equation}
\end{itemize}

The overall objective encourages that, for clean samples, the similarity between the image features and the positive prompt exceeds that of the negative prompt, while the opposite holds for noisy labels. By jointly optimizing \(\mathcal{L}_1\) and \(\mathcal{L}_2\), the model aligns visual and textual embeddings while explicitly modeling pseudo-label uncertainty. The total loss is written as:
\begin{equation}
    \mathcal{L} = \mathcal{L}_1 + \lambda \mathcal{L}_2,
\end{equation}
where \(\lambda\) balances the contribution of the negative prompt supervision.

\subsubsection{Parameter-Efficient Fine-Tuning of the Visual Encoder}

On the visual side, we perform parameter-efficient fine-tuning of the visual encoder to adapt CLIP to task-specific visual distributions. Concretely, we adopt Visual Prompt Tuning (VPT) \cite{jia2022visual}, where a small number of learnable visual prompt tokens are prepended to the input sequence of the visual transformer. Only these prompts are optimized, while the pre-trained backbone parameters remain frozen. This design allows efficient adaptation and is further regularized by the negative prompt learning objective, which supervises the visual prompts from a complementary “negative discrimination” perspective, improving robustness against limited supervision.

\begin{table*}[htbp]
\centering
\small
\setlength{\tabcolsep}{1pt}
\resizebox{1.0\textwidth}{!}{
\begin{tabular}{
>{\centering\arraybackslash}p{2.0cm}  % Dataset
>{\centering\arraybackslash}p{0.8cm}  % B(%)
*{8}{>{\centering\arraybackslash}p{1.5cm}}
>{\centering\arraybackslash\columncolor[HTML]{E6E6E6}}p{1.5cm}
>{\centering\arraybackslash\columncolor[HTML]{E6E6E6}}p{1.5cm}
}
\toprule
Dataset & $B(\%)$ & \textbf{Zero-shot} & \textbf{Random} & \textbf{Entropy} & \textbf{CoreSet} & \textbf{BADGE} & \textbf{ALFA-Mix} & \textbf{GCNAL} & \textbf{CEC} & \textbf{Ours}$_{CoOp}$ & \textbf{Ours} \\
\midrule

\multirow{3}{*}{DTD} & 1 & \multirow{3}{*}{44.3} & 38.4$\pm$0.2 & 35.2$\pm$0.8 & -- & 40.2$\pm$5.0 & -- & 38.8$\pm$0.9 & 47.9$\pm$1.2 & 48.1$\pm$1.0 & \textbf{52.0$\pm$0.8} \\
    & 2 &      & 44.2$\pm$2.9 & 40.9$\pm$2.0 & 44.9$\pm$0.9 & 46.9$\pm$1.4 & 49.6$\pm$0.4 & 44.8$\pm$0.4 & 52.8$\pm$1.0 & 53.0$\pm$0.9 & \textbf{55.0$\pm$0.7} \\
    & 5 &      & 54.1$\pm$2.9 & 49.3$\pm$1.4 & 52.9$\pm$2.3 & 56.2$\pm$1.1 & 55.6$\pm$1.9 & 55.0$\pm$1.4 & 58.2$\pm$2.0 & 58.8$\pm$1.1 & \textbf{61.8$\pm$0.9} \\

\midrule
\multirow{3}{*}{Caltech101} & 1 & \multirow{3}{*}{91.3} & 88.2$\pm$3.4 & 86.1$\pm$4.6 & -- & 88.2$\pm$1.7 & -- & 88.4$\pm$3.3 & 88.7$\pm$1.5 & 92.4$\pm$1.6 & \textbf{93.9$\pm$0.4} \\
           & 2 &      & 88.4$\pm$3.2 & 89.3$\pm$1.4 & 91.1$\pm$2.0 & 89.8$\pm$2.7 & 89.7$\pm$0.8 & 89.8$\pm$2.0 & 89.0$\pm$1.3 & 93.9$\pm$1.5 & \textbf{94.4$\pm$0.4} \\
           & 5 &      & 91.1$\pm$1.1 & 89.4$\pm$0.8 & 91.3$\pm$0.3 & 92.2$\pm$0.1 & 92.3$\pm$0.3 & 92.4$\pm$0.9 & 92.8$\pm$0.5 & 94.1$\pm$0.4 & \textbf{95.4$\pm$0.3} \\

\midrule
\multirow{3}{*}{EuroSAT} & 1 & \multirow{3}{*}{42.0} & 82.2$\pm$1.0 & 70.5$\pm$2.0 & -- & 80.6$\pm$0.7 & -- & 82.1$\pm$1.4 & 82.8$\pm$1.6 & 84.5$\pm$0.9 & \textbf{91.2$\pm$0.6} \\
        & 2 &      & 86.1$\pm$1.0 & 78.1$\pm$4.3 & 84.5$\pm$1.5 & 85.6$\pm$0.7 & 86.1$\pm$0.3 & 84.0$\pm$0.9 & 86.2$\pm$0.6 & 86.8$\pm$1.2 & \textbf{91.4$\pm$0.5} \\
        & 5 &      & 87.8$\pm$0.6 & 84.8$\pm$2.1 & 87.9$\pm$1.4 & 87.5$\pm$0.5 & 88.3$\pm$0.3 & 88.2$\pm$0.6 & 88.0$\pm$0.9 & 90.0$\pm$0.5 & \textbf{93.6$\pm$0.6} \\

\midrule
\multirow{3}{*}{FGVC-Aircraft} & 1 & \multirow{3}{*}{24.9} & 18.4$\pm$0.6 & 19.7$\pm$1.1 & -- & 17.8$\pm$1.7 & -- & 18.4$\pm$0.6 & 20.3$\pm$1.1 & 21.2$\pm$1.2 & \textbf{27.2$\pm$1.0} \\
              & 2 &      & 21.2$\pm$1.4 & 22.0$\pm$2.1 & 19.7$\pm$1.4 & 20.7$\pm$1.1 & 20.2$\pm$1.5 & 21.2$\pm$0.4 & 22.3$\pm$1.0 & 23.5$\pm$1.1 & \textbf{27.5$\pm$0.9} \\
              & 5 &      & 26.0$\pm$1.0 & 24.7$\pm$0.6 & 23.0$\pm$0.1 & 25.7$\pm$1.0 & 28.7$\pm$0.4 & 26.3$\pm$0.7 & 27.1$\pm$0.3 & 27.8$\pm$0.4 & \textbf{29.7$\pm$0.8} \\

\midrule
\multirow{3}{*}{Flowers102} & 1 & \multirow{3}{*}{67.3} & 60.2$\pm$2.2 & 55.2$\pm$4.7 & -- & 53.5$\pm$5.3 & -- & 60.2$\pm$2.3 & 64.1$\pm$2.4 & 66.1$\pm$1.9 & \textbf{74.5$\pm$0.8} \\
           & 2 &      & 66.3$\pm$2.2 & 65.4$\pm$3.5 & 62.2$\pm$0.7 & 68.2$\pm$1.7 & 74.0$\pm$0.8 & 64.0$\pm$4.0 & 75.6$\pm$2.5 & 76.8$\pm$2.9 & \textbf{80.9$\pm$0.7} \\
           & 5 &      & 82.6$\pm$2.2 & 80.3$\pm$2.8 & 76.2$\pm$2.6 & 86.2$\pm$1.6 & 88.7$\pm$1.2 & 80.0$\pm$2.9 & 88.2$\pm$1.6 & 88.9$\pm$0.8 & \textbf{89.9$\pm$0.5} \\

\midrule
\multirow{3}{*}{UCF101} & 1 & \multirow{3}{*}{64.3} & 55.4$\pm$2.7 & 53.1$\pm$3.9 & -- & 50.7$\pm$3.0 & -- & 55.3$\pm$3.7 & 57.6$\pm$1.8 & 60.8$\pm$1.2 & \textbf{75.4$\pm$0.9} \\
       & 2 &      & 66.6$\pm$1.2 & 62.1$\pm$1.9 & 65.4$\pm$1.4 & 63.7$\pm$1.5 & 66.9$\pm$1.5 & 63.9$\pm$1.9 & 67.0$\pm$0.8 & 68.5$\pm$1.5 & \textbf{75.8$\pm$0.7} \\
       & 5 &      & 73.8$\pm$0.4 & 72.8$\pm$0.7 & 74.1$\pm$3.0 & 75.3$\pm$1.2 & 75.4$\pm$1.9 & 73.1$\pm$1.4 & 76.2$\pm$0.6 & 77.0$\pm$0.6 & \textbf{77.9$\pm$0.5} \\

\midrule
\multirow{3}{*}{Average} & 1 & \multirow{3}{*}{55.7} & 57.1 & 53.3 & -- & 55.2 & -- & 57.2 & 60.2 & 62.6 (\textcolor{blue}{+2.4}) & \textbf{69.1} (\textcolor{blue}{+8.9}) \\
        & 2 &      & 62.1 & 59.6 & 61.3 & 62.5 & 64.4 & 61.3 & 65.5 & 66.7 (\textcolor{blue}{+1.2}) & \textbf{70.8} (\textcolor{blue}{+5.3}) \\
        & 5 &      & 69.2 & 66.9 & 67.6 & 70.5 & 71.5 & 69.2 & 71.8 & 72.8 (\textcolor{blue}{+1.0}) & \textbf{74.7} (\textcolor{blue}{+2.9}) \\

\bottomrule
\end{tabular}}
\caption{Accuracy comparison of different AL methods for CLIP adaptation under varying annotation budgets with CoOp. Best results are in bold.}
\label{tab:al_results}
\end{table*}

\begin{table*}[t]
\centering
\small
\setlength{\tabcolsep}{4pt}
\begin{tabular}{
>{\centering\arraybackslash}p{2.0cm}  % Dataset
>{\centering\arraybackslash}p{0.8cm}  % B(%)
*{6}{>{\centering\arraybackslash}p{1.5cm}}
>{\centering\arraybackslash\columncolor[HTML]{E6E6E6}}p{1.5cm}
>{\centering\arraybackslash\columncolor[HTML]{E6E6E6}}p{1.5cm}
}
\toprule
Dataset & $B(\%)$ & \textbf{Zero-shot} & \textbf{Random} & \textbf{Entropy} & \textbf{CoreSet} & \textbf{BADGE} & \textbf{CEC} & \textbf{Ours}$_{VPT}$ & \textbf{Ours} \\
\midrule

\multirow{3}{*}{DTD} & 1 & \multirow{3}{*}{44.3} & 38.9$\pm$2.9 & 39.3$\pm$1.5 & -- & 42.5$\pm$2.7 & 45.9$\pm$1.0 & 48.2$\pm$1.0 & \textbf{52.0$\pm$0.8} \\
    & 2 &      & 49.3$\pm$3.1 & 44.4$\pm$1.3 & 47.0$\pm$1.0 & 50.0$\pm$0.8 & 52.8$\pm$0.9 & 53.4$\pm$0.9 & \textbf{55.0$\pm$0.7} \\
    & 5 &      & 58.2$\pm$2.5 & 57.0$\pm$1.7 & 56.6$\pm$0.7 & 59.7$\pm$1.9 & 61.4$\pm$0.4 & 61.0$\pm$0.7 & \textbf{61.8$\pm$0.9} \\

\midrule
\multirow{3}{*}{Flowers102} & 1 & \multirow{3}{*}{67.3} & 61.0$\pm$3.7 & 55.3$\pm$4.2 & -- & 56.7$\pm$4.3 & 64.6$\pm$5.2 & 65.8$\pm$1.1 & \textbf{74.5$\pm$0.8} \\
           & 2 &      & 70.6$\pm$1.4 & 66.8$\pm$3.8 & 65.9$\pm$1.4 & 70.8$\pm$2.1 & 75.9$\pm$1.3 & 76.6$\pm$0.9 & \textbf{80.9$\pm$0.7} \\
           & 5 &      & 80.5$\pm$1.5 & 81.6$\pm$1.2 & 79.6$\pm$2.1 & 84.8$\pm$0.7 & 85.7$\pm$0.9 & 86.7$\pm$0.6 & \textbf{89.9$\pm$0.5} \\

\midrule
\multirow{3}{*}{UCF101} & 1 & \multirow{3}{*}{64.3} & 58.8$\pm$2.7 & 60.0$\pm$4.8 & -- & 61.3$\pm$3.8 & 63.7$\pm$1.2 & 65.6$\pm$1.2 & \textbf{75.4$\pm$0.9} \\
       & 2 &      & 66.5$\pm$0.8 & 67.4$\pm$1.5 & 67.6$\pm$1.9 & 68.6$\pm$0.7 & 70.8$\pm$0.9 & 71.8$\pm$0.8 & \textbf{75.8$\pm$0.7} \\
       & 5 &      & 75.0$\pm$0.7 & 74.0$\pm$1.5 & 75.2$\pm$1.2 & 77.9$\pm$0.4 & 78.0$\pm$0.9 & 77.5$\pm$0.6 & \textbf{77.9$\pm$0.5} \\

\midrule
\multirow{3}{*}{Average} & 1 & \multirow{3}{*}{58.6} & 52.9 & 51.5 & -- & 53.5 & 58.1 & 59.9 (\textcolor{blue}{+1.8}) & \textbf{67.3} (\textcolor{blue}{+9.2})\\
        & 2 &      & 62.1 & 59.5 & 60.2 & 63.1 & 66.5 & 67.3 (\textcolor{blue}{+0.8}) & \textbf{70.6} (\textcolor{blue}{+4.1}) \\
        & 5 &      & 71.2 & 70.9 & 70.5 & 74.1 & 75.0 & 75.1 (\textcolor{blue}{+0.1}) & \textbf{76.5} (\textcolor{blue}{+1.5}) \\

\bottomrule
\end{tabular}
\caption{Accuracy comparison of different AL methods for CLIP adaptation under varying annotation budgets with VPT. Best results are in bold.}
\label{tab:al_results_vpt}
\end{table*}

\begin{table*}[htbp]
\centering
\small
\setlength{\tabcolsep}{4pt}
\begin{tabular}{
>{\centering\arraybackslash}p{2.0cm}  % Dataset
>{\centering\arraybackslash}p{0.8cm}  % B(%)
*{6}{>{\centering\arraybackslash}p{1.5cm}}
>{\centering\arraybackslash\columncolor[HTML]{E6E6E6}}p{1.5cm}
>{\centering\arraybackslash\columncolor[HTML]{E6E6E6}}p{1.5cm}
}
\toprule
Dataset & $B(\%)$ & \textbf{Zero-shot} & \textbf{Random} & \textbf{Entropy} & \textbf{CoreSet} & \textbf{BADGE} & \textbf{CEC} & \textbf{Ours}$_{MaPLe}$ & \textbf{Ours} \\
\midrule

\multirow{3}{*}{DTD} & 1 & \multirow{3}{*}{44.3} & 37.1$\pm$3.5 & 36.0$\pm$7.8 & -- & 38.3$\pm$0.7 & 45.8$\pm$2.2 & 46.9$\pm$1.0 & \textbf{52.0$\pm$0.8} \\
    & 2 &      & 46.0$\pm$1.8 & 37.8$\pm$6.5 & 40.7$\pm$0.4 & 45.4$\pm$1.7 & 50.8$\pm$2.0 & 52.3$\pm$0.9 & \textbf{55.0$\pm$0.7} \\
    & 5 &      & 55.7$\pm$2.0 & 54.9$\pm$2.2 & 52.0$\pm$1.2 & 57.7$\pm$1.3 & 56.7$\pm$0.9 & 57.8$\pm$0.8 & \textbf{61.8$\pm$0.9} \\

\midrule
\multirow{3}{*}{Flowers102} & 1 & \multirow{3}{*}{67.3} & 62.2$\pm$1.4 & 61.2$\pm$2.5 & -- & 67.0$\pm$3.1 & 66.0$\pm$2.3 & 68.5$\pm$1.1 & \textbf{74.5$\pm$0.8} \\
           & 2 &      & 72.0$\pm$2.7 & 64.0$\pm$4.2 & 70.0$\pm$4.4 & 70.1$\pm$4.1 & 77.2$\pm$1.9 & 77.8$\pm$0.9 & \textbf{80.9$\pm$0.7} \\
           & 5 &      & 82.5$\pm$2.5 & 83.0$\pm$2.1 & 79.0$\pm$2.6 & 86.2$\pm$0.2 & 86.5$\pm$0.4 & 87.4$\pm$0.6 & \textbf{89.9$\pm$0.5} \\

\midrule
\multirow{3}{*}{UCF101} & 1 & \multirow{3}{*}{64.3} & 64.4$\pm$3.9 & 59.7$\pm$3.2 & -- & 62.0$\pm$0.7 & 65.8$\pm$1.6 & 68.3$\pm$1.0 & \textbf{75.4$\pm$0.9} \\
       & 2 &      & 69.6$\pm$1.3 & 66.2$\pm$3.1 & 69.8$\pm$0.1 & 68.8$\pm$3.1 & 70.9$\pm$2.4 & 71.5$\pm$0.8 & \textbf{75.8$\pm$0.7} \\
       & 5 &      & 76.9$\pm$0.8 & 73.8$\pm$2.4 & 76.3$\pm$0.5 & 76.6$\pm$0.8 & 77.6$\pm$0.2 & 77.8$\pm$0.6 & \textbf{77.9$\pm$0.5} \\

\midrule
\multirow{3}{*}{Average} & 1 & \multirow{3}{*}{58.6} & 54.6 & 52.3 & -- & 55.8 & 59.2 & 61.2 (\textcolor{blue}{+2.0}) & \textbf{67.3} (\textcolor{blue}{+8.1}) \\
        & 2 &      & 62.5 & 56.0 & 60.2 & 61.4 & 66.3 & 67.2 (\textcolor{blue}{+0.9}) & \textbf{70.6} (\textcolor{blue}{+4.3}) \\
        & 5 &      & 71.7 & 70.6 & 69.1 & 73.5 & 73.6 & 74.3 (\textcolor{blue}{+0.7}) & \textbf{76.5} (\textcolor{blue}{+2.9}) \\

\bottomrule
\end{tabular}
\caption{Accuracy comparison of different AL methods for CLIP adaptation under varying annotation budgets with MaPLe. Best results are in bold.}
\label{tab:al_results_maple}
\end{table*}

\begin{table*}[t]
\centering
\small
\setlength{\tabcolsep}{4pt}
\begin{tabular}{
>{\centering\arraybackslash}p{2.0cm}  % Dataset
>{\centering\arraybackslash}p{0.8cm}  % B(%)
*{6}{>{\centering\arraybackslash}p{1.5cm}}
>{\centering\arraybackslash\columncolor[HTML]{E6E6E6}}p{1.5cm}
>{\centering\arraybackslash\columncolor[HTML]{E6E6E6}}p{1.5cm}
}
\toprule
Dataset & $B(\%)$ & \textbf{Zero-shot} & \textbf{Random} & \textbf{Entropy} & \textbf{CoreSet} & \textbf{BADGE} & \textbf{CEC} & \textbf{Ours}$_{CoOp}$ & \textbf{Ours} \\
\midrule

\multirow{3}{*}{DTD} & 1 & \multirow{3}{*}{53.0} & 45.7$\pm$0.9 & 40.0$\pm$2.8 & -- & 45.1$\pm$2.5 & 55.1$\pm$2.6 & 56.8$\pm$1.1 & \textbf{58.3$\pm$1.0} \\
    & 2 &      & 50.0$\pm$1.9 & 43.8$\pm$2.0 & 51.8$\pm$1.9 & 52.1$\pm$1.1 & 56.4$\pm$0.8 & 57.9$\pm$0.9 & \textbf{59.6$\pm$0.8} \\
    & 5 &      & 60.4$\pm$0.7 & 56.9$\pm$2.0 & 59.7$\pm$1.9 & 61.7$\pm$2.1 & 63.7$\pm$1.7 & 65.2$\pm$2.1 & \textbf{67.0$\pm$1.8} \\

\midrule
\multirow{3}{*}{Flowers102} & 1 & \multirow{3}{*}{79.3} & 75.0$\pm$1.3 & 76.2$\pm$0.9 & -- & 72.6$\pm$3.8 & 80.0$\pm$1.4 & 81.2$\pm$1.0 & \textbf{83.0$\pm$0.7} \\
           & 2 &      & 77.1$\pm$1.7 & 73.2$\pm$2.1 & 74.1$\pm$2.1 & 79.1$\pm$4.0 & 86.4$\pm$2.0 & 87.8$\pm$2.2 & \textbf{89.5$\pm$1.5} \\
           & 5 &      & 87.1$\pm$2.4 & 87.8$\pm$2.0 & 84.2$\pm$2.2 & 92.8$\pm$0.2 & 93.6$\pm$0.7 & 94.3$\pm$1.2 & \textbf{95.8$\pm$0.9} \\

\midrule
\multirow{3}{*}{UCF101} & 1 & \multirow{3}{*}{74.2} & 73.3$\pm$1.6 & 72.4$\pm$1.0 & -- & 73.7$\pm$1.2 & 74.9$\pm$1.3 & 76.1$\pm$1.3 & \textbf{77.9$\pm$1.1} \\
       & 2 &      & 76.4$\pm$0.7 & 72.3$\pm$1.4 & 74.1$\pm$1.5 & 75.1$\pm$2.6 & 77.5$\pm$0.7 & 78.8$\pm$0.8 & \textbf{80.6$\pm$0.6} \\
       & 5 &      & 82.4$\pm$1.7 & 80.0$\pm$0.8 & 80.5$\pm$0.3 & 82.0$\pm$0.9 & 82.5$\pm$1.5 & 83.9$\pm$1.0 & \textbf{85.7$\pm$0.8} \\

\midrule
\multirow{3}{*}{Average} & 1 & \multirow{3}{*}{68.8} & 64.7 & 62.9 & -- & 63.8 & 70.0 & 71.4 (\textcolor{blue}{+1.4}) & \textbf{73.1} (\textcolor{blue}{+3.1}) \\
        & 2 &      & 67.8 & 63.1 & 66.7 & 68.8 & 73.4 & 74.8 (\textcolor{blue}{+1.4}) & \textbf{76.6} (\textcolor{blue}{+3.2}) \\
        & 5 &      & 76.6 & 74.9 & 74.8 & 78.8 & 79.9 & 81.1 (\textcolor{blue}{+1.2}) & \textbf{82.8} (\textcolor{blue}{+2.9})  \\

\bottomrule
\end{tabular}
\caption{Accuracy comparison of different active learning methods for CLIP adaptation under ViT-L/14 backbone using CoOp. Best results are in bold.}
\label{tab:al_results_vitl14}
\end{table*}

\subsection{Uncertainty-Driven AL Framework}

We integrate our dual-prompt CLIP model into a round-based AL pipeline, in which the model is re-initialized at the beginning of each round to prevent error accumulation.

Before the first AL round, no manually labeled samples are available. We perform zero-shot inference on the entire unlabeled set \(\mathcal{U}\) using pre-trained CLIP. For each class, we select the top-\(K\) samples with the highest predicted probabilities to form the initial sampled unlabeled set \(S_U^{(0)}\).

At round \(r\), the model is trained using both:
\begin{itemize}
    \item the current sampled labeled set \(S_L^{(r)}\) with obtained human annotated labels, and
    \item a pseudo-labeled set \(S_U^{(r)}\) sampled from the unlabeled data according to the clean probability.
\end{itemize}

After training, the adapted model assigns pseudo-labels \(\hat{y}\) to all remaining unlabeled samples and computes \(p^{\text{clean}}_{\hat{y}}\) for each sample according to Eq.\ref{pclean}. Samples are grouped by their pseudo-label class and ranked within each group.

\textbf{Uncertainty-Based Query Selection.} To select samples for human annotation, we identify the most uncertain samples within each class (i.e., those with the lowest \(p^{\text{clean}}_{\hat{y}}\)). Given a total budget \(B\) per round and \(C\) classes, we compute the per-class selection number as \(\lfloor B/C \rfloor\). Any remainder \((B \bmod C)\) is allocated by picking the most uncertain samples across classes in descending order of uncertainty. This ensures approximate class balance while fully utilizing the annotation budget.

\textbf{Confident Sample Mining.} For each pseudo-label class, we select the top-\(k\) samples with the highest \(p^{\text{clean}}_{\hat{y}}\) values as most confident samples. These samples are incorporated into the training set for the next round.

The above procedure repeats for multiple AL rounds. At the beginning of each round, the model is re-initialized and retrained on the updated \(S_L^{(r)}\) and \(S_U^{(r)}\) sets, preventing confirmation bias and accumulation of pseudo-label errors.

\section{Experiments}
\subsection{Experimental Setup}

Our AL experimental setting basically follows \cite{safaei2025active}. In all experiments, we perform 6 rounds of AL, where at each round we select $1\%$ of the entire unlabeled data for human annotation. For AL methods that require initial labeled data we perform random sampling in the first round. For a fair comparison, we conduct each experiment across three different seeds and report the mean and standard deviations of the results.

\subsubsection{Datasets and Baselines}
Following previous methods \cite{safaei2025active}, we use six publicly available image classification datasets including Caltech101 \cite{fei2004learning}, DTD \cite{cimpoi2014describing}, EuroSAT \cite{helber2019eurosat}, FGVCAircraft \cite{maji2013fine}, Flowers102 \cite{nilsback2008automated} and UCF101 \cite{soomro2012ucf101}. These datasets cover a variety of different visual classification tasks, from general to fine-grained objects such as texture classification, constituting a comprehensive benchmark.

We compare our method against a suite of state-of-the-art AL approaches, namely, \textbf{ALFA-Mix} \cite{parvaneh2022active}, \textbf{BADGE} \cite{ash2019deep}, \textbf{CEC} \cite{safaei2025active}, \textbf{CoreSet} \cite{sener2017active}, \textbf{Entropy} \cite{wang2014new}, \textbf{GCNAL} \cite{caramalau2021sequential} and \textbf{Random}.

\subsubsection{Implementation Details}
Without specific declaration, we use CLIP ViT-B/16 as our backbone for all comparing methods. We use SGD as the optimizer, and the learning rate is chosen from $lr=\{0.001, 0.01, 0.03\}$, along with a cosine decay scheduler. The number of total training epochs is selected from \{10, 20\}. The training batch-size of sampled labeled set and unlabeled set are set to 16 and 64, respectively. For both positive and negative in dual-prompt learning, the number of tunable context tokens is $M=16$, with the class token positioned at the end, and the tunable tokens for VPT is $M_v=20$. We conduct all the experiments on one NVIDIA RTX 3090 GPU.

\subsection{Main Results}

Table~\ref{tab:al_results} summarizes the performance of different active learning methods for adapting CLIP under limited annotation budgets. In all experiments, we follow a unified evaluation protocol to ensure a fair comparison among different active learning strategies. Specifically, each active learning method is used solely to select informative samples for annotation under a given budget ($1\%$, $2\%$, or $5\%$ of the unlabeled data). The selected samples, together with their ground-truth labels, are then used to fine-tune a pre-trained CLIP model using CoOp~\cite{zhou2022learning}, a representative and widely adopted parameter-efficient prompt learning method for CLIP. Under this setting, the final classification accuracy directly reflects the quality of the samples selected by each active learning strategy, rather than differences in downstream adaptation techniques.

In Table~\ref{tab:al_results}, we report two variants of our method. \textbf{Ours}$_{CoOp}$ denotes the performance obtained by fine-tuning CLIP with CoOp using the samples selected by our proposed active learning strategy, which follows exactly the same evaluation protocol as all other baselines. \textbf{Ours} corresponds to the test accuracy of our full framework after each active learning round, where model training is performed jointly with the sample selection process.

From the quantitative results, our method consistently outperforms all competing active learning baselines across all datasets and annotation budgets. Under the extremely low annotation budget of $1\%$, \textbf{Ours}$_{CoOp}$ already achieves clear improvements over existing methods, with an average accuracy of $62.6\%$, surpassing the strongest baseline CEC ($60.2\%$) by $+2.4\%$. When evaluating the full framework, our method further boosts the average performance to $69.1\%$, yielding a substantial gain of $+8.9\%$ over CEC. Similar trends are observed under $2\%$ and $5\%$ annotation budgets, where our full framework achieves average accuracies of $70.8\%$ and $74.7\%$, respectively, consistently outperforming all baseline methods. Notably, the performance gap is particularly pronounced on challenging datasets such as EuroSAT, UCF101, and Flowers102, demonstrating the robustness of our uncertainty-driven sample selection strategy across diverse visual domains.

The additional performance gain of our full framework over \textbf{Ours}$_{CoOp}$ mainly stems from the fact that our framework leverages unlabeled data during the active learning process, providing auxiliary supervision beyond the selected labeled samples.

\subsubsection{Results with VPT and MaPLe}
Table~\ref{tab:al_results_vpt} and Table~\ref{tab:al_results_maple} report the performance of different AL strategies for CLIP adaptation using VPT and MaPLe under various annotation budgets. In each AL round, each comparing method, including Random, Entropy, CoreSet, BADGE, and CEC, is used to select informative samples, which are then used to fine-tune CLIP with the same PEFT method. This setup ensures a fair comparison of sample selection quality, as all methods share the same adaptation procedure.

\textbf{Ours}$_{VPT}$ denotes the performance obtained by fine-tuning VPT using the samples selected by our proposed method. As shown in Table~\ref{tab:al_results_vpt}, \textbf{Ours}$_{CoOp}$ consistently outperforms all baseline methods across datasets and annotation budgets. For example, under $1\%$, $2\%$ and $5\%$ budgets, \textbf{Ours}$_{VPT}$ surpasses the strongest comparative method CEC by +1.8\%, +0.8\% and +0.1\%, respectively, confirming that our sample selection strategy effectively identifies informative samples for VPT-based adaptation. In addition, \textbf{Ours}, representing the full framework where sample selection and model training are performed jointly, achieves even higher accuracies.

Similar trends are observed under experiments using MaPLe. As shown in Table~\ref{tab:al_results_maple}, our method surpasses the strongest comparative method by +2.0\%, +0.9\% and +0.7\%, respectively. It can be observed from the experiments that combining our method with VPT or MaPLe does not fully unleash its potential. Our full framework achieves superior performance, yielding improvements of +8.1\%, +4.3\%, and +2.9\% over the strongest baseline under $1\%$, $2\%$ and $5\%$ annotation budgets, respectively.

\subsection{Results under Other Backbone}
Table~\ref{tab:al_results_vitl14} presents the performance of different active learning methods on CLIP ViT-L/14. From the results, it is evident that our method consistently outperforms all competing baselines across all datasets and annotation budgets. For example, under the extremely low annotation budget of $1\%$, our method achieves an average accuracy of $71.4\%$, surpassing the strongest baseline CEC ($70.0\%$) by $+1.4\%$. When evaluating the full framework, \textbf{Ours} further boosts the average performance to $73.1\%$, yielding an additional gain of $+3.1\%$. Similar trends are observed under larger annotation budgets, where the full framework achieves average accuracies of $76.6\%$ and $82.8\%$, respectively.

These results demonstrate that our method is not only effective under the commonly used ViT-B/16 backbone but also generalizes well to a larger ViT-L/14 backbone, highlighting the robustness and adaptability of our approach across different backbone architectures.

\section{Conclusion}
This paper proposes a dual-prompt tuning framework for explicit uncertainty modeling in active CLIP adaptation. By introducing positive and negative learnable prompts in CLIP’s textual branch, our method robustly estimates pseudo-label reliability, enabling effective uncertainty-aware sample selection and confident pseudo-label mining. Extensive experiments across six datasets, three PEFT paradigms (CoOp, VPT, MaPLe), and two backbones (ViT-B/16, ViT-L/14) demonstrate that our approach consistently outperforms state-of-the-art active learning methods under limited annotation budgets.

% \section*{Acknowledgments}

% The preparation of these instructions and the \LaTeX{} and Bib\TeX{}
% files that implement them was supported by Schlumberger Palo Alto
% Research, AT\&T Bell Laboratories, and Morgan Kaufmann Publishers.
% Preparation of the Microsoft Word file was supported by IJCAI.  An
% early version of this document was created by Shirley Jowell and Peter
% F. Patel-Schneider.  It was subsequently modified by Jennifer
% Ballentine, Thomas Dean, Bernhard Nebel, Daniel Pagenstecher,
% Kurt Steinkraus, Toby Walsh, Carles Sierra, Marc Pujol-Gonzalez,
% Francisco Cruz-Mencia and Edith Elkind.

\bibliographystyle{named}
\bibliography{ijcai26}

\end{document}